\documentclass[11pt,a4paper]{article}
\usepackage[hyperref]{emnlp2020}
\usepackage{times}
\usepackage{latexsym}

\usepackage{microtype}
\usepackage[super]{nth}
\usepackage{bm}
\usepackage{graphicx}
\usepackage{multicol}
\usepackage{float}
\usepackage{dblfloatfix}
\usepackage[utf8x]{inputenc} 

\DeclareUnicodeCharacter{8294}{-}
\DeclareUnicodeCharacter{12540}{-}

\aclfinalcopy 
\def\aclpaperid{68} 

\title{NutCracker at WNUT-2020 Task 2: Robustly Identifying Informative COVID-19 Tweets using Ensembling and Adversarial Training}

\author{
\\
  \\
  \And
  Priyanshu Kumar and Aadarsh Singh \\
  \\
  Indian Institute of Technology (Indian School of Mines) Dhanbad, India \\
  {\tt \{kpriyanshu256, aadarshsingh191198 \}@gmail.com} \\
  \And
  \\
  \\
  \\}

\date{}

\begin{document}
\maketitle
\begin{abstract}
We experiment with COVID-Twitter-BERT and RoBERTa models to identify informative COVID-19 tweets. We further experiment with adversarial training to make our models robust. The ensemble of COVID-Twitter-BERT and RoBERTa obtains a F1-score of 0.9096 (on the positive class) on the test data of WNUT-2020 Task 2 and ranks \nth{1} on the leaderboard. The ensemble of the models trained using adversarial training also produces similar result.
\end{abstract}

\section{Introduction}

Since 2006, Twitter has been a popular social network where people express their thoughts and opinions on various topics. The enforcement of lockdown in various parts of the world, due to COVID-19, led to an increased usage of social media. Thus, the world has witnessed a plethora of tweets since the beginning of COVID-19. People have tweeted about many issues regarding the pandemic, mostly about the increasing infection rate, carelessness and incapability of governance and authorities to handle the increasing rate of cases. In addition, various preventive measures have also been conveyed through tweets. 

Although there have been more than 600 million English tweets on COVID-19 \citep{lamsal2020corona} , only a few of them are informative enough to be used by various monitoring systems to update their databases. Manual identification of these informative tweets can be tedious and erroneous. Hence, there is a dire need to develop systems in the form of machine learning models that can help us in filtering informative tweets.

In this paper, we present our approaches for the shared task “Identification of informative COVID-19 English Tweets” organized under the Workshop on Noisy User-generated Text (W-NUT). Our method makes use of ensembles consisting of Bidirectional Encoder Representations from Transformers (BERT) \citep{devlin2018bert} pretrained on COVID-19 tweets \citep{muller2020covid} and Robustly Optimized BERT Pretraining Approach (RoBERTa) \citep{liu2019roberta}. We also experiment with adversarial training so as to create models that generalise well and are robust.

The rest of the paper is organized as follows: Related work has been discussed in Section \ref{sec:relwork}, followed by a brief description of the data used in Section \ref{sec:data}. The proposed methods and experimental settings \footnote{Source code available at \url{https://github.com/kpriyanshu256/WNUT-2020-Task-2}} have been elaborated in Section \ref{sec:process}, \ref{sec:methods}  and \ref{sec:exp_set}. Section \ref{sec:results} and \ref{sec:error} contains the results and error analysis respectively. Section \ref{sec:end} concludes the paper and also includes possible future work.

\section{Related Work}
\label{sec:relwork}

There has been much research to identify informative tweets during times of emergency and disaster. \citet{neppalli2018deep} explored the performance of traditional machine learning algorithms and deep learning approaches for identifying informative tweets during disaster. They created manual features using the content of tweets for machine learning approaches and also tried out features obtained from Convolutional and Recurrent networks for deep learning approaches. A multi-modal approach for classifying informative tweets during disaster was proposed by \citet{madichetty2020classifying}. They combined the features obtained from text by a Convolutional Neural Network and the VGG-16 features obtained from the image accompanying the tweet, using late fusion for better performance than models using only text or only image.

\citet{roy2020classification} proposed a classification and summarisation approach to identify informative tweets during the Fani cyclone (which occurred in 2019, affecting large parts of South-East Asia). They trained a Support Vector Machine (SVM) on linguistic features and Parts of Speech (POS) tags from tweets to identify informative tweets. To summarise the informative tweets, they experimented with Latent Semantic Analysis (LSA) and Luhn summarisation techniques. \citet{zahera2019fine} experimented with the transformer based architecture BERT to classify disaster related tweets into multi-label information types. They preprocessed the tweets of TREC-IS dataset before feeding them into BERT to produce significantly better results than the median score. In addition, they also experimented with Focal loss instead of binary cross-entropy loss.

A new dataset for identifying informative tweet was released by \citet{aggarwal2019classification}. The results of various machine learning algorithms using GloVe \citep{pennington2014glove} word embeddings, syntactic information in the form of tf-idf vectors and BERT embeddings were also presented in the work. 

Due to the abundance of tweets related to COVID-19, many works have been done to analyze the content, intent and effect of such tweets. \citet{singh2020first} presented an analysis of COVID-19 tweets on the grounds of location, content and misinformation spread. A comprehensive study about the content of misinformative COVID-19 tweets and other aspects associated with them  have been done by \citet{shahi2020exploratory}.


\section{Dataset}
\label{sec:data}

The dataset \citep{covid19tweet} provided to the participants of the shared task contains 10,000 English COVID-19 tweets, out of which 4719 are labeled as INFORMATIVE and 5281 are labeled as UNINFORMATIVE. The tweets were annotated by 3 independent annotators and an inter-annotator agreement score of Fleiss' Kappa at 0.818 was obtained. The dataset contains the tweet ID, the tweet and the corresponding label.


\section{Data Preprocessing}
\label{sec:process}

Twitter data contains a lot of noise. Therefore, preprocessing on Twitter data will help the pretrained models in better performance. We perform the following data preprocessing steps, most of which have been inspired from \citet{muller2020covid} :

\begin{enumerate}
\item Unescape HTML tags
\vspace{-0.2cm} \item Remove unnecessary spaces, tabs and newlines
\vspace{-0.2cm} \item Replacing the the mentioned hyperlinks in the tweets (depicted as HTTPURL), with URL. A simple explanation for this could be that “URL” is a more commonly used expression of hyperlinks than HTTPURL.

\vspace{-0.2cm} \item Using the Python emoji \footnote{\url{https://pypi.org/project/emoji/}} library to demojise the emojis i.e. replace them with a short textual description.
\end{enumerate}

The user handles were already replaced by @USER in the tweets, hence no processing was required.

\section{Models}
\label{sec:methods}

We experiment with the following models and techniques:

\begin{enumerate}

\item {\bf COVID-Twitter-BERT}: BERT is based on the Transformer architecture \cite{vaswani2017attention}. It consists of multi-attention heads which apply a sequence-to-sequence transformation on the input text sequence. 
For its training, BERT makes use of the following objectives:
(a) learn to predict a masked token using the left and right context of the text sequence (Masked Language Model)
(b) learn to predict whether two sentences occur in continuation or not (Next Sentence Prediction)

\citet{muller2020covid} pretrain the large version of BERT (BERT-Large) on COVID-19 related tweets posted by users between January 12 and April 16, 2020. This version of BERT has a better understanding of the given data as compared to BERT-large, which is pretrained on texts from Wikipedia. Hence, COVID-Twitter-BERT will be even more beneficial for the task when fine tuned.

\item {\bf RoBERTa Large}: RoBERTa has the same architecture as BERT, but is different from BERT on the grounds of pretraining, which helps in better optimisation and performance. RoBERTa is pretrained on a larger dataset as compared to BERT, uses a larger batch size and replaces the Next Sentence Prediction objective. It also uses dynamic masking pattern as a better alternative to the static masking pattern used in BERT, i.e. RoBERTa duplicates the data and masks those differently each time, whereas BERT will mask the data only once.

\item {\bf Adversarial Training}: With time, adversarial training is gaining popularity in Natural Language Processing (NLP) as well. In the field of Computer Vision, adversarial training is done by perturbing the input images slightly and minimising the adversarial loss. In NLP, the nature of input being discrete, small perturbations are done on the word embeddings. Adversarial training not only increases the robustness of models but also helps in better generalisation. Both properties are beneficial and desirable for identification of informative tweets.

Although many approaches for adversarial training in NLP have been developed, we experiment with the approach proposed by \citet{miyato2016adversarial} with a slight modification. In their approach, first the word embeddings are normalized. The gradients are then computed using the data and the required perturbations are created using the obtained gradients.

Let the sequence of (normalized) word embedding vectors of a text be \bm{$t$}. The model parameters are represented by \bm{$\theta$}. The probability of the text belonging to class $y$ is given by $p(y|\bm{t;\theta})$. The adversarial perturbations $\bm{z}_{adv}$ are computed as follows: 

\vspace{-0.05cm}

\[
\bm{g} = \nabla _t\: log\: p(y|\bm{t;\theta})
\]
\vspace{-0.95cm}

\[
\bm{z}_{adv} = -\epsilon \bm{g} /\!\parallel \bm{g}\! \parallel _2
\]

where $\epsilon$ is a hyper-parameter controlling the size of the perturbations. The adversarial loss is defined as :
\vspace{-0.05cm}
\[
    \bm{L}_{adv}(\bm{\theta}) = - \frac{1}{N} \sum_{n=1}^{N} log\: p(y_n | \bm{t}_n + \bm{z}_{adv,n};\bm{\theta})
\]

By using the gradients calculated from the above loss, the weights of the model are updated (the non-perturbed word embeddings of the model are updated). The slight modification in our experiments is that we do not normalize our pretrained word embedding of the model, since it might change the semantic meaning of the pretrained word embeddings. We perform adversarial training on both COVID-Twitter-BERT and RoBERTa Large models using $\epsilon=1$ .

\item {\bf Ensembling}: The ensembling of the predictions for our submissions is done at two levels- 
\begin{enumerate}
    \item Fold level i.e. the predictions obtained by the models trained using the different folds (during cross validation) are averaged.
    
    \item Model level i.e. the fold level averaged predictions of the two different models are ensembled using averaging.
 
\end{enumerate}

\end{enumerate}

\section{Experimental Settings}
\label{sec:exp_set}
We concatenate the training and the validation data and perform a 5-fold stratified cross validation to train our models. Each fold is trained for 5 epochs using early stopping with patience of 3 and tolerance of 1e-3. The models are optimised using AdamW \citep{DBLP:journals/corr/abs-1711-05101} with a learning rate of 2e-5 and a batch size of 16. The models have been implemented using Pytorch \citep{paszke2019pytorch} and Huggingface's Transformers \citep{wolf2019transformers} library.

\section{Results}
\label{sec:results}

We evaluate the performance of the models and their ensemble using cross-validation (CV). We also tabulate the models' performance on the test set as evaluated by the F1 score on the positive class.

The ensembling done at two levels increases the robustness of the results. The out-of-folds predictions were used to find the optimal threshold of the submissions, 0.498 for the ensemble without adversarial training and 0.487 for the ensemble with adversarial training. We also compare our results with the fastText-baseline \citep{joulin2016bag} by the organizers. Table \ref{tab:results} shows the results of our experiments.

\begin{table*}[bp]
\renewcommand\thetable{2}
\centering
\begin{tabular}{|l|c|}
\hline \hspace{5.5cm} \textbf{Tweet} & \textbf{Label} \\ \hline
\begin{minipage}{0.75\textwidth}
\vspace{0.15cm}
Election Judge Hospitalized After Primary Dies Of Coronavirus \#RIP \#SemperFi fought for his life while ⁦@USER got her hair done \#LetThatSinkIn “face of Chicago” thinks she’s more important than Chicagoans welfare \#LightfootLiedPeopleDied HTTPURL
\vspace{0.15cm}
\end{minipage} & UNINFORMATIVE\\

\hline
\begin{minipage}{0.75\textwidth}
\vspace{0.15cm}
Austin area nursing home residents who test positive for COVID-19 but do not need to be in the hospital will soon be moving to one of two new “isolation facilities," one in Travis County, one in Williamson County: HTTPURL @USER
\vspace{0.01cm}
\end{minipage} & INFORMATIVE \\ \hline

\begin{minipage}{0.75\textwidth}
\vspace{0.15cm}
LOCAL NEWS SHOUTOUT: Have a family member or close friend with a Michigan connection who has died from COVID-19 and would like to share their story with @USER Please contact Georgea Kovanis at gkovanis@USER
\vspace{0.15cm} 
\end{minipage} & INFORMATIVE \\ \hline

\begin{minipage}{0.75\textwidth}
\vspace{0.15cm}
BREAKING: Southern AB \#coronavirus case rumored to be Steve Busey, the younger brother of movie star @USER According to our sources, Steve works in the oil \&amp; gas industry and is a huge @USER fan. \#COVIDー19 \includegraphics[height=2\fontcharht\font`\B]{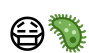} HTTPURL 
\vspace{0.15cm} 
\end{minipage} & UNINFORMATIVE \\ \hline

\end{tabular}
\caption{\label{tab:topk} Some common highly misclassified samples.}
\end{table*}

\begin{table}[ht]
\renewcommand\thetable{1}
\centering
\begin{tabular}{|p{4.3cm}|c|c|}
\hline \hspace{1.6cm} \textbf{Model} & \textbf{CV} & \textbf{Test} \\ \hline
Baseline - fastText & - & 0.7503 \\ \hline
COVID-Twitter-BERT & 0.9622 & - \\ \hline
RoBERTa Large & 0.9560 & - \\ \hline
COVID-Twitter-BERT Adv. & 0.9632 & -\\ \hline
RoBERTa Large Adv. & 0.9578 & - \\ \hline
COVID-Twitter-BERT + RoBERTa Large & 0.9636 & {\bf 0.9096} \\ \hline
COVID-Twitter-BERT Adv. + RoBERTa Large Adv. & {\bf 0.9655} & 0.9082 \\ \hline
\end{tabular}
\caption{\label{tab:results} Comparison of results.}
\end{table}

The ensemble of COVID-Twitter-BERT and RoBERTa Large performs the best on the test data and also obtains the \nth{1} rank on the leaderboard. Owing to its pretraining, COVID-Twitter-BERT performs better than RoBERTa Large. The adversarial training of the models is also found to boost the scores. The ensemble of adversarial models produces results similar to the ensemble of models trained without adversarial training.

\section{Analysis}
\label{sec:error}

     

We perform our analysis on the out-of-folds predictions from our two ensembles - without adversarial training and with adversarial training.




We calculate the binary cross entropy loss for all samples and then examine some of the top common mis-classifications by both the ensembles ( Table \ref{tab:topk}).

We observe that there are instances where our models are mistaken. A possible reason might be the variance in gold-labels of samples from annotator-to-annotator. This acts as noise in the labels of the data which is used to train the normal ensemble models. On the other hand, adversarial training adds some noise to the samples. Hence, the models in the adversarial ensemble have been trained on data which has a combination of existing noise and externally added noise. We believe that because of this difference in noise, the models in the two ensembles must be behaving in different ways. To inspect this, we examine the number of samples which have been inferred incorrectly by one ensemble and correctly by the other.

We observe that the normal ensemble misclassified a total of 277 samples whereas, only 260 samples were misclassified by the adversarial ensemble. Moreover, out of the 277 examples that were misclassified by the normal ensemble, 102 were correctly predicted by the adversarial ensemble. On the other hand, only 85 out of the 260 samples misclassified by the adversarial ensemble, were correctly predicted by the normal ensemble. Thus, it is evident that the models in both the ensembles have learnt different patterns from the data. 



\section{Conclusion}
\label{sec:end}

We explored the performance of COVID-Twitter-BERT and RoBERTa-Large at identifying COVID-19 English tweets that are informative in nature. Their ensemble achieves the state-of-the-art performance. Adversarial training is found to improve our model further. For future work, we can pretrain other Transformer-based models on COVID-19 tweets. Data augmentation techniques can help us generate more data for training models.  We can also experiment with combining models trained with and without adversarial training. 

\section*{Acknowledgments}
We thank Google Colab for providing free GPU services for experimentation purposes.


\bibliographystyle{acl_natbib}
\bibliography{anthology,emnlp2020}


\end{document}